\documentclass[10pt,journal,compsoc]{IEEEtran}

\ifCLASSOPTIONcompsoc
  \usepackage[nocompress]{cite}
\else
  \usepackage{cite}
\fi

\usepackage{graphicx}
\usepackage{capt-of}
\usepackage{xcolor}

\renewcommand{\textcolor}[2]{#2}
\renewcommand{\color}[1]{}

\usepackage{caption}
\color{black}

\usepackage{amsmath,amsfonts,dsfont,pifont,bm,bbm,mathrsfs,mathtools,nicefrac}
\usepackage{algpseudocode,listings}
\usepackage{booktabs,multirow,adjustbox,diagbox,threeparttable}
\usepackage{ulem}
\definecolor{citeblue}{RGB}{48,111,186}
\usepackage[hidelinks]{hyperref}

\usepackage{placeins}
\usepackage{flafter}  

\usepackage[capitalize]{cleveref} 
\crefname{section}{Sec.}{Secs.}
\Crefname{section}{Section}{Sections}
\crefname{table}{Tab.}{Tabs.}
\Crefname{table}{Table}{Tables}
\crefname{figure}{Fig.}{Figs.}
\Crefname{figure}{Figure}{Figures}
\crefname{equation}{Eq.}{Eqs.}
\Crefname{equation}{Equation}{Equations}
\newcommand{\tocite}[1]{\textcolor{red}{[TO CITE]}}

\ifCLASSINFOpdf
\else
\fi

\usepackage{graphicx} 
\usepackage{float} 
\usepackage{subfigure} 
\usepackage{graphicx}
\usepackage{etoolbox}
\usepackage{caption} \newcommand{\insertfig}

\makeatletter
\apptocmd{\@maketitle}{\centering\insertfig}{}{}
\makeatother

\newcommand{\sysname}{SERES}

\begin{document}
%
\title{\huge{\sysname: Semantic-Aware Neural Reconstruction \\ from Sparse Views}}
%
%
%

\author{Bo~Xu$^*$, 
       Yuhu~Guo$^*$, 
       Yuchao~Wang$^*$, 
      Wenting~Wang, 
       Yeung~Yam, 
      Charlie~C.L.~Wang$^\dag$,
        and~Xinyi~Le$^\dag$
\thanks{ $^*$~~denotes  equal contribution to this work.}
\thanks{ $^\dag$~ denotes the corresponding authors.\\Bo Xu, Yuchao Wang and Xinyi  Le are with the Department of Automation, Shanghai Jiao Tong University, Shanghai, China (email:
~lexinyi@sjtu.edu.cn). \\Yuhu Guo and  Charlie~C.L.~Wang are with Department of Mechanical and Aerospace Engineering, The University of Manchester, Manchester, UK (email:~charlie.wang@manchester.ac.uk). \\
Wenting~Wang and Yeung~Yam are with Department of Mechanical and Automation Engineering, The Chinese University of Hong Kong, Hong Kong SAR.}}

\IEEEtitleabstractindextext{%
\begin{abstract}
We propose a semantic-aware neural reconstruction method to generate 3D high-fidelity models from sparse images. To tackle the challenge of severe radiance ambiguity caused by mismatched features in sparse input, we enrich neural implicit representations by adding patch-based semantic logits that are optimized together with the signed distance field and the radiance field. A novel regularization based on the geometric primitive masks is introduced to mitigate shape ambiguity. The performance of our approach has been verified in experimental evaluation. The average chamfer distances of our reconstruction on the DTU dataset can be reduced by 44\% for SparseNeuS and 20\% for VolRecon. When working as a plugin for those dense reconstruction baselines such as NeuS and Neuralangelo, the average error on the DTU dataset can be reduced by 69\% and 68\% respectively.

\end{abstract}

}


\maketitle

\newcommand{\rev}[2]{{#2}}
\newcommand{\yuchao}[1]{\textcolor[rgb]{0.8,0.20,0.70}{\textit{[Yuchao:~#1]}}}

\newcommand{\yh}[1]{\textcolor[rgb]{0.4,0.10,0.80}{\textit{[Yuhu:~#1]}}}
\newcommand{\wt}[1]{\textcolor{red}{\textit{[Wenting:~#1]}}}
\newcommand{\xy}[1]{\textcolor{green}{\textit{[Le:~#1]}}}

\newcommand{\charlie}[1]{\textbf{\textcolor[rgb]{0.60,0.00,0.00}{[Charlie::~#1]}}}
\newcommand{\Xinyi}[1]{\textbf{\textcolor[rgb]{0.90,0.20,0.80}{[Xinyi::~#1]}}}

\IEEEdisplaynontitleabstractindextext

%
\IEEEpeerreviewmaketitle

\IEEEraisesectionheading{
}

\vspace{-5mm}
\section{Introduction}\label{sec:intro}


%
Neural implicit representation-based 3D reconstruction methods~\cite{NeuRIS, wang2021neus} have gained significant popularity in recent years due to reconstruct geometric structures with remarkable fidelity.

However, the construction of accurate and intricate geometry poses a challenge for these methods when the input comprises only a sparse set of images.  
%
The challenge is mainly caused by the limited coverage of views, which prevents the effective matching on many pixels in the input views.
As a result, this leads to shape-ambiguity~\cite{kaizhang2020,wei2021nerfingmvs,Zhu_2023_CVPR} in the radiance field to be learned.

To tackle this challenge, scene priors are usually embedded into radiance field learning through a specially designed framework that integrates neural networks\cite{yu2020pixelnerf}~\cite{long2022sparseneus} with the capability to acquire prior scene knowledge or utilize special 3D geometric data structures\cite{sparesmvsnerf}.
An alternative paradigm is based on incorporating external geometric cues\cite{yu2022monosdf} (e.g., surface normals\cite{NeuRIS}, depth maps\cite{wang2023sparsenerf,Ren_2023_CVPR}, and albedo information\cite{NeRFactor}) as supplementary constraints to enhance the precision of scene geometry. 
However, both paradigms involve significant costs -- i.e., introducing additional neural networks during training requires substantial computational overhead, and incorporating geometric cues necessitates acquiring extra data.
An inherently pertinent question arises: 
\textit{How can we add a low-cost scene prior into neural implicit representations effectively?} 

\begin{figure*}[htbp]
    \centering
    \includegraphics[width=\linewidth]{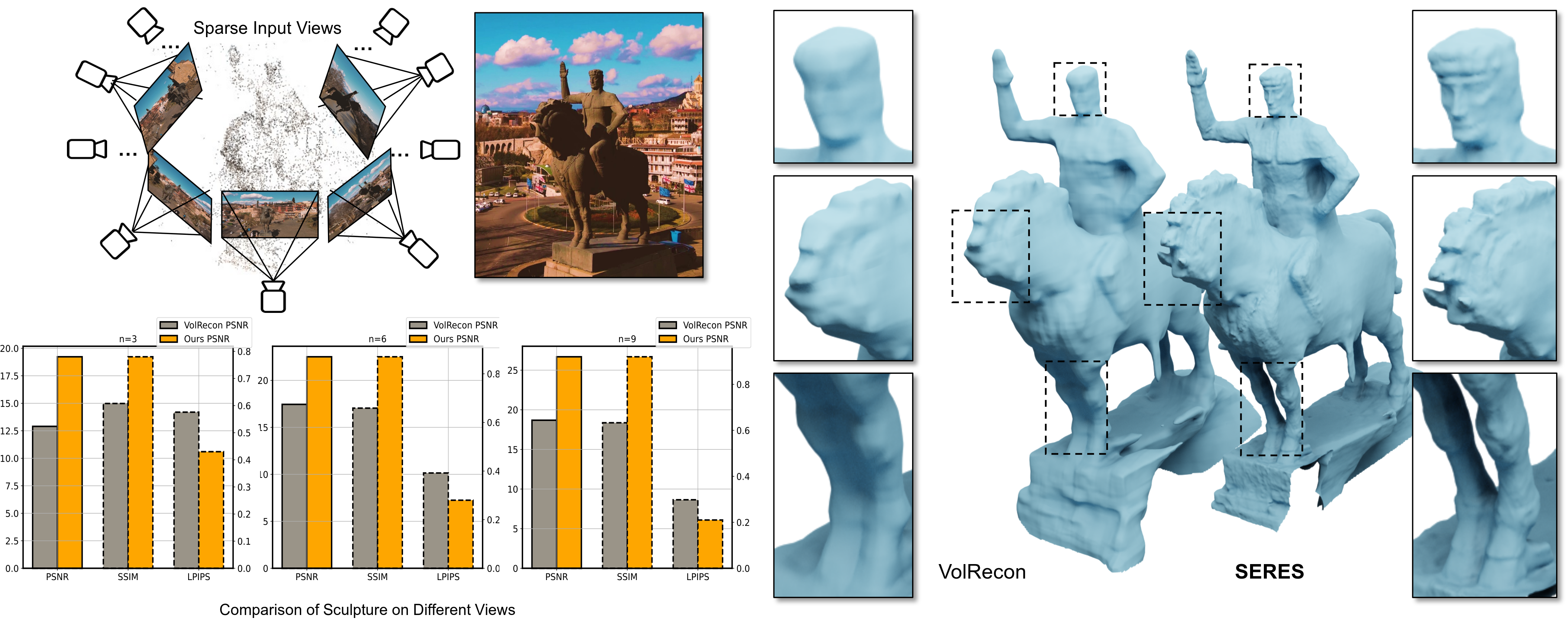}
    \caption{We argue that the quality of 3D models from sparse views can be significantly improved by incorporating semantic awareness into the neural reconstruction pipeline. When reconstructing the statue of Vakhtang I Gorgasali from \textit{nine} views (five are illustrated in the figure), our \sysname{} approach can generate more geometric details than the state-of-the-art VolRecon~\cite{Ren_2023_CVPR}. We report novel view synthesis metrics for 3, 6, and 9  viewpoints input (center random selection), shown in the bottom left. Solid bars refer to the left y-axis, and dashed bars to the right y-axis.}
    \label{fig:figure_1_asia_update}
\end{figure*}

Our idea was inspired by the engineering drawing practice, where the engineers can rely on only \textit{three} views to accurately reconstruct the 3D geometry of most objects. 
The key concept of this involves dividing the object into common semantic patches in various views, thereby acquiring a prior understanding of the scene.
It can be inferred that incorporating a prior based on semantic patch matching would be a beneficial choice for 3D reconstruction from sparse views.
%

Based on this insight, this paper proposes a novel \textbf{SE}mantic-aware framework for \textbf{RE}construction from \textbf{S}pare views (\sysname).
%
%
Specifically, we enrich the neural field representation by semantic logits, the initial values of which can be obtained by a training-free segmentation model and a vision transformer. Reliable feature matching and therefore high-fidelity reconstruction can be achieved by optimizing these semantic logits together with the signed distance field and the radiance field. During optimization, the geometric primitive masks are also employed as regularization, which offers additional constraints to alleviate shape-ambiguity. 
%
As shown in Fig.\ref{fig:figure_1_asia_update}, our \sysname{} can successfully reconstruct intricate sculptures using only nine viewpoints, while preserving the correct geometric structure and capturing fine details.

In summary, we make the following contributions:
\begin{itemize}
\item We present a novel semantic field which embed patch-based semantic priors into neural implicit representations. This prior provides semantic-aware feature matching for improving the fidelity of sparse view reconstruction.


\item We propose a point prompt-guided regularization technique, mitigating the shape ambiguity of reconstructed shape and improving the completeness of geometric structures. 


\item This technique has shown its generality to benefit different neural implicit representation frameworks, serving as an effective plugin \textcolor{blue}{with acceptable computational overhead.}
\end{itemize}

The performance of our \sysname{} has been tested in a variety of academic benchmarks and real-world scenarios.

\section{Related Work}\label{sec:related-work}
\subsection{Sparse Multi-view Stereo}
Modeling 3D surfaces from sparse views is a difficult task, as accurately aligning infrequent regions with 3D space presents a challenge~\cite{Kutulakos2000}. Previous attempts~\cite{yu2020pixelnerf, wang2023sparsenerf} aim to reduce the number of required viewpoints, but they do not extract geometries. Instead, they focus mainly on novel view synthesis~\cite{Niemeyer2021Regnerf, sparesmvsnerf} or estimating unknown camera positions~\cite{sinha2023sparsepose}. Recent studies such as \cite{long2022sparseneus, yu2022monosdf,Ren_2023_CVPR} have made positive strides in sparse view reconstruction. However, previous research has not been successful in creating high-quality and easy-to-use sparse view reconstructions. SC-NeuS~\cite{Huang_Zou_Zhang_Cao_Shan_2024} requires multi-view constraints to refine camera poses. MonoSDF~\cite{yu2022monosdf} necessitates precise depth and normal information, while SparseNeuS~\cite{long2022sparseneus} often fails to reconstruct complex shapes, leaving them with holes in the surface. VolRecon~\cite{Ren_2023_CVPR} can produce satisfactory results, but tends to smooth textures in detail-rich regions and generate noise around them. 



%

\subsection{Neural Implicit Surface Reconstruction}
NeRF~\cite{mildenhall2020nerf} has revolutionized the field of 3D world understanding by introducing novel view synthesis. This technique shares similarities with reconstruction, yet originates from a distinct conceptual foundation. Then, NeuS\cite{wang2021neus} highlights the  difficulty  of depicting high-quality surfaces in learned implicit representations, attributed to a lack of surface constraints. Directly estimating signed distance function~(SDF) from multi-view images\cite{zhang2021learning} in neural implicit representations can lead to geometric biases in surface reconstruction\cite{yariv2021volume, yu2022monosdf}. To obtain precise SDF representation, recent efforts aim to incorporate additional geometric constraints~\cite{Darmon_2022_CVPR,huang2024neusurf} or cues~\cite{yu2022monosdf}, including object masks~\cite{yariv2020multiview,Niemeyer_2020}, normal priors~\cite{NeuRIS,verbin2022refnerf}, depth information~\cite{Bian_2023_CVPR}, viewpoint selection~\cite{na2024uforecon} and photometric consistency~\cite{fu2022geoneus}.

Notably, previous techniques have disregarded the significance of semantic coherence across multiple view, especially in cases of sparse views where semantics considerably assist in recuperating geometry. 
%
    
%

\subsection{SAM for 3D understanding} 
The Segment Anything Model (SAM)~\cite{Kirillov_2023_ICCV} is highly proficient in carrying out detailed image segmentation. However, it lacks 3D understanding, which is currently undergoing active review.   For example, SAD~\cite{sad} employs SAM for the segmentation of rendered depth maps. This method uses cues with embedded geometry to map semantic segmentation into 3D via depth for stereoscopic display. Further, Liu et al.\cite{liu2023segment} made the first attempt to employ SAM models for self-supervised learning in large-scale 3D point clouds. In addition, SAM3D\cite{Yang2023SAM3DSA} presents a simple but effective framework for segmenting 3D scenes. This framework projects SAM segmentation masks from RGB frames onto 3D point clouds, then iteratively fuses them using 3D registration to produce  scene segmentation. Similarly, Zhang et al.\cite{Zhang2023SAM3DZ3} apply SAM for segmenting Bird's Eye View maps, using its masks for zero-shot 3D object detection.
\color{blue}
Moving beyond simple projection, Panoptic Lifting \cite{Siddiqui2023panoptic} learns a neural field that directly outputs 3D-consistent panoptic labels by volumetrically rendering semantic and instance logits, demonstrating a more integrated approach to 3D semantic understanding. 
\color{black}
Perhaps  the most relavant network to our study is Anything-3D\cite{Shen2023Anything3DTS}. The pipeline merges visual-language model\cite{Rombach_2022_CVPR} and SAM for modeling, offering a plausible single-view 3D reconstruction system. 

Basically, previous efforts on utilizing SAM for 3D understanding have been focused primarily on extending 2D SAM to 3D segmentation~\cite{cen2023segment,Yang2023SAM3DSA,liu2023segment,Zhang_2024_WACV} or simply using SAM as a tool for image matting~\cite{Shen2023Anything3DTS,Liu2023One2345}. However, the potential for utilizing SAM's semantic patches as constraints to aid in multi-view reconstruction or SDF estimation tasks remains largely unexplored.

\section{Method}

Our goal is to reconstruct high-quality geometric structures from \textit{sparse} multi-view inputs of a object via semantic matching encoding and geometric primitive regularization.
Firstly, leveraging the visual foundation model SAM~\cite{Kirillov_2023_ICCV}, we obtained patches and primitives from input views without any annotations in \ref{sec:sam}.
Next, we will explain how to utilize the obtained patches to generate accurate semantic matching features in \ref{sec:semantic matching}.
Then, we utilize the semantic matching features to construct a semantic-aware neural field in \ref{sec:neural implicit model}.
Finally, by overlaying the supervision of geometry primitive masks and the RGB+geometry loss of the neural field, we ensure the correct learning of the neural field \ref{sec:loss}.

\subsection{Generating Patches and Primitives}~\label{sec:sam}
\begin{figure}[!ht]
\includegraphics[width=1.0\linewidth]{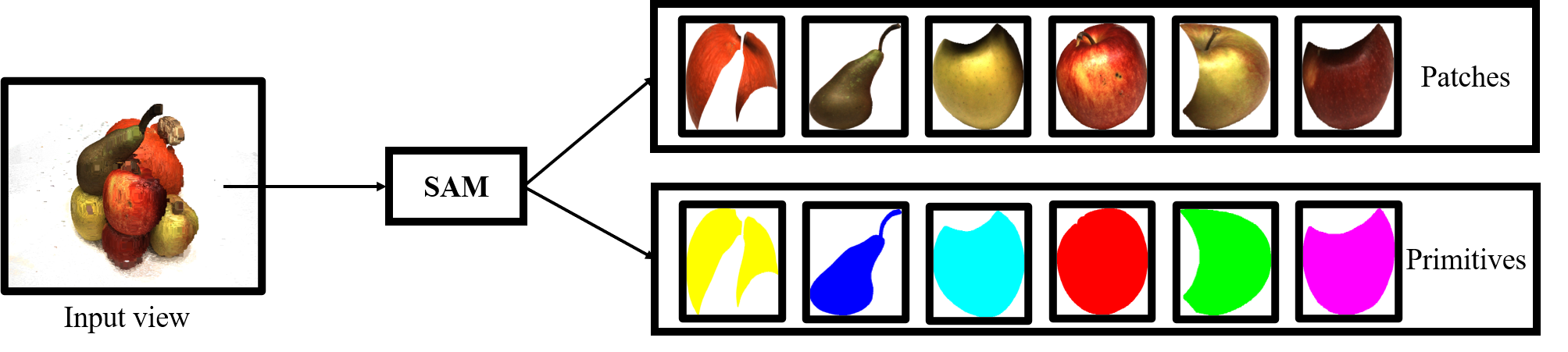}
\vspace{-15pt}
\caption{
How to generate patches and primitives from the input view image via the segment anything model.
}\label{fig:sam-generation}
\vspace{-10pt}
\end{figure}

\begin{figure*}[t]
    \centering
     \includegraphics[width=1.0\linewidth]{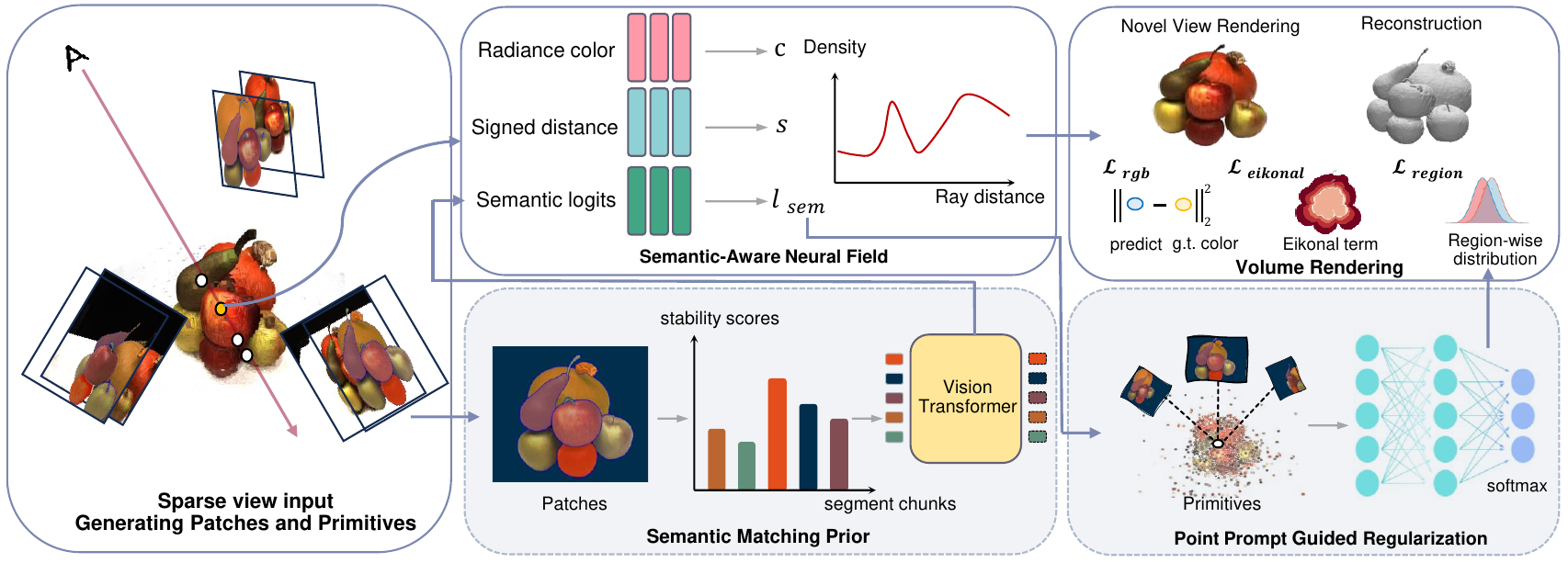}
    \vspace{-15pt}
    \caption{
The overall pipeline encompasses the capability to reconstruct high-quality geometric meshes from sparse view input.
We generate patches and primitives from sparse input and construct a semantic-aware neural field.
The semantic matching prior for patches and the  point prompt guided regularization are the fundamental components of the semantic-aware neural field.
We optimize the learning of the neural field by incorporating the RGB loss obtained from volume rendering, along with the eikonal loss and the region loss obtained from point prompt guided regularization.
}
    \label{fig:overview_with_sam}
    \vspace{-5pt}
\end{figure*}
Given a collection of $M$ sparse input images $\left\{I_{1}, \cdots, I_{M}\right\}$ of the target object, we utilize the SAM model~\cite{Kirillov_2023_ICCV} to generate semantic patches and geometric primitives.
For any single image input view $I_{i}$ of the $\left\{I_{1}, \cdots, I_{M}\right\}$, we utilize SAM to extract $N$ patches $\left\{P^{pat}_{1}(i), \cdots, P^{pat}_{N}(i)\right\}$ and corresponding $N$ primitives $\left\{P^{pri}_{1}(i), \cdots, P^{pri}_{N}(i)\right\}$ in an unsupervised zero-shot manner as demonstrated in Fig.~\ref{fig:sam-generation}.
Here we consistently select the top N patches and primitive pairs based on the stability score, where $N$ is a hyperparameter we set, and stability is the confidence-like score output of the original SAM.

The $j$-th patch $P^{pat}_{j}(i) \in 3\times H \times W $ is the RGB image of a single, independent semantic object, while the $j$-th primitive $P^{pri}_{j}(i)\in C\times H \times W$ corresponds to the semantic label of the patch $P^{pat}_{j}(i)$. 
Patches $\left\{P^{pat}_{1}(i), \cdots, P^{pat}_{N}(i)\right\}$ are utilized to construct semantic matching features, which are incorporated into the training process of the neural field.
Primitives $\left\{P^{pri}_{1}(i), \cdots, P^{pri}_{N}(i)\right\}$ are utilized to implement as the semantic supervision, enforcing the neural field to learn appropriate representations under sparse input conditions.
$C$ is a hyper-parameter which denotes the number of categories.
How $\left\{P^{pat}_{1}(i), \cdots, P^{pat}_{N}(i)\right\}$ and $\left\{P^{pri}_{1}(i), \cdots, P^{pri}_{N}(i)\right\}$ are integrated into our framework is demonstrated in figure~\ref{fig:overview_with_sam}.
$\left\{P^{pat}_{1}(i), \cdots, P^{pat}_{N}(i)\right\}$ will be utilized to construct semantic matching prior in Sec.~\ref{sec:semantic matching}.
$\left\{P^{pri}_{1}(i), \cdots, P^{pri}_{N}(i)\right\}$ will be used as the one of the supervision in Sec.~\ref{sec:loss}.

\subsection{Semantic Matching Prior}~\label{sec:semantic matching}
To obtain precise semantic matching prior, we have devised a resilient mechanism to acquire them:
For the sparse input set $\left\{I_{1}, \cdots, I_{N}\right\}$, given the semantic patch set $\left\{P^{pat}_{1}(i), \cdots, P^{pat}_{N}(i)\right\}$  for the $i-th$ view $I_{i}$, this module will construct a semantic matching prior feature $\hat{F}_{i}$ for $i-th$ view.
Initially, we extract features from all patches {$\left\{P^{pat}_{1}(i), \cdots, P^{pat}_{N}(i)\right\}$} of view $I_{i}$ (generated from \ref{sec:sam}) using a pre-trained ViT model.
The weights of the ViT model are kept frozen, ensuring that it remains unaltered during the training process.
Subsequently, an embedding with a shape of $1 \times n$ is obtained for each patch. 
%
%
Through a concatenation operation, we derive a feature vector $f_{i}$ of size ${m} \times n$ as the semantic patch embedding for each input view $I_{i}$.
After applying identical operations to all $\left\{I_{1}, \cdots, I_{N}\right\}$, we acquire a collection of view-specific semantic patch features: $\left\{f_{1}, \cdots, f_{N}\right\}$ for the input view $1\sim N$.



%
The feature set $\left\{f_{1}, \cdots, f_{N}\right\}$ contains only the feature information from its own view, while the matching information is determined by both its own view and the features of other views. 
Therefore, we employ cross attention operations to aggregate features between different view $i$ and view $j$ as equation~\ref{eq:multi}.
\begin{equation}\label{eq:multi}
\begin{aligned}
    \hat{f}_{i}^{j} & =  MultiHead(f_{j}, f_{i}, f_{i}) \\
                & = Concat(head_{1},\cdots, head_{M}) 
\end{aligned}
\end{equation} 
where $M$ is set to be $8$ and $head_{i}$ is defined as:
\begin{equation}\label{eq:attn}
    \begin{aligned}
        head_{i}(f_{j}, f_{i}, f_{i}) = softmax(\frac{f_{j}f_{i}^{T}}{\sqrt{d_k}})f_{i}
    \end{aligned}
\end{equation}
where $\sqrt{d_k}$ is a pre-defined scaling factor. 
By this operation, we introduce attentative guidance based on the similarity of other views' features for each views, thus incorporating semantic matching prior.


The guidance via attention operation of the feature from view $I_j$ to view $I_i$ is denoted as $\hat{f}_{i}^{j}$.
Similarly, the guided attentive feature encoding of the remaining input views with respect to $I_i$ can be obtained: $\left\{\hat{f}_{i}^{1}, \cdots, \hat{f}_{i}^{N}\right\}$. 
Finally, a shared multi-layer perceptron (MLP) is utilized to aggregate this set of attentive features, mapping the feature dimension from $(N-1)\times m \times n$ to $m \times n$, resulting in $\hat{F}_{i}$.
\begin{equation}
    \hat{F}_{i}=MLP(\left\{\hat{f}_{i}^{1}, \cdots, \hat{f}_{i}^{N}\right\})
\end{equation}

Through the utilization of feature similarity across various view point, the corresponding feature undergoes fusion to produce a resilient semantic matching prior feature $\hat{F}_{i}$.
We incorporate this feature $\hat{F}_{i}$ into the radiance field $f_{c}^{s}$ of the neural implicit model via a naive MLP as equation ~\ref{eq:semlogit}.

\subsection{Semantic-Aware Neural Field}~\label{sec:neural implicit model} 
The typical neural implicit representation employs calibrated view images to implicitly capture an object's surface and appearance using a signed distance field \(f_{sdf}(\mathbf{x})\rightarrow s\) and a geometry-aware radiance field \(f_{c}(\mathbf{x}, \mathbf{v}, s)\rightarrow \sigma,c \), where \( \mathbf{x} \) denotes a 3D position, $s$ is the signed distance function value,  and \( \mathbf{v} \in S^2 \) is the $2D$ viewing direction encoding.

To effectively incorporate semantic matching priors and maximize the utilization of semantic matching information, we define our semantic-aware neural field $f_{c}^{s}$ as equation \ref{eq:semlogit}: 

\begin{equation}\label{eq:semlogit}
    f_{c}^{s}(x, s, \hat{F}_{i})\rightarrow \sigma, c, l_{sem}
\end{equation}
where the view-specific semantic matching prior for view $i$ is denoted as $\hat{F}_{i}$.
$l_{sem}$ is defined as the logit of the class distribution for the spatial point $x$.
The additional output $l_{sem}$ undergoes processing through the softmax function and is then mapped to predetermined logits that represent semantic distributions. 
These logits are utilized to predict the class labels associated with each pixel, allowing for the optimization of semantic learning within the neural field. 
The logits will be further explored in section \ref{sec:loss}.
The prior for semantic matching, denoted as $\hat{F}_{i}$, operates as an additional input feature that strengthens the neural field's capacity to explicitly recognize corresponding elements. 
This will be further demonstrated in Section \ref{sec:semantic matching}.

\subsection{Loss Function}\label{sec:loss}

We optimize the following overall loss function:
\begin{equation}
    L = L_{rgb} + \lambda_{1}L_{region} + \lambda_{2}L_{eikonal}
\end{equation}
$L_{eikonal}$ is proposed in ~\cite{gropp2020implicit} to add a $\mathcal{L}_{2}$ regularization which encourage the learned surface to be smooth.
Give $n$ sparse RGB images input $ I=\left\{I_{1}, I_{2}, \cdots, I_{N}, \right\}$, $L_{rgb}$ is defined as:
\begin{equation}
    L_{rgb} = \frac{1}{N}\sum_{i=1}^{N}\sum_{p}||I_{i}(p)-\hat{I}_{i}(p)||_{2}^{2}
\end{equation}
where $p$ denotes the pixel in the image $I_{i}$, and $\hat{I}_{i}(.)$ can be queried via volumetric rendering with the $\sigma$ and $c$ from ~\ref{eq:semlogit}.
$\lambda_{1}$ and $\lambda_{2}$, as the balance factor corresponding to each term, are set to $0.1$ and $0.05$ respectively.

$L_{region}$ is defined as the semantic-aware region regularization loss.
%
\color{blue}
Specifically, for each input image $i$, we render the 3D semantic logits $l_{sem}$(from Eq. \ref{eq:semlogit}) from its corresponding camera viewpoint using differentiable volumetric rendering~\cite{mildenhall2020nerf}. This process yields a 2D semantic mask $M_{sem}(i)$
 that is geometrically aligned with the input image.
\color{black}
To this end, we can define a region semantic regularization loss $L_{region}$:
%


\begin{equation}\color{blue}\label{eq:region_semantic}
L_{region} =\sum_{i=1}L_{CE}(\left\{P^{pri}_{1}(i), \cdots, P^{pri}_{N}(i)\right\}, M_{sem}(i))
\end{equation}
Where $L_{CE}(., .)$ denotes the cross-entropy loss, $\left\{P^{pri}_{1}(i), \cdots, P^{pri}_{N}(i)\right\}$ is defined in \ref{sec:sam}.
%

%

%
The $L_{region}$ term utilizes region-level semantic information from foundation model SAM to enhance the distinction of intricate object boundaries within the scene. 
Through the ablation study, we discovered the $L_{region}$'s capability to effectively eliminate noise within the mesh.
This enables our semantic-aware neural field to obtain improved representations in sparse view reconstruction.

\color{blue}
\section{Experiment setup}\label{sec:exp}

In this section, we provide a comprehensive description of our experimental setup. We detail the datasets used in~\cref{sec:exp:dataset}, the baseline methods and competitors in~\cref{sec:exp:baseline}, and our implementation details in~\cref{sec:exp:implement}.
\color{black}

\subsection{Dataset}~\label{sec:exp:dataset} 
We trained our framework, \sysname, using both the DTU~\cite{jensen2014large} dataset for comparison with previous studies~\cite{wang2021neus,long2022sparseneus}, 
and the BlendedMVS~\cite{yao2020blendedmvs} dataset for verifying complex scenes. 
The scenes from the DTU~\cite{jensen2014large} dataset exhibit diverse materials, appearances, and geometries, thereby creating challenging scenarios for reconstruction algorithms. Each scene consists of either 49 or 64 images with an image resolution of $1600 \times 1200$.  Our selection of scenes from the BlendedMVS dataset includes images ranging from $31$ to $143$, along with a resolution of $768$ $\times$ $576$ and corresponding masks.

Since the \sysname{} is specifically designed for sparse inputs, we selected three viewpoints for evaluation in each scene to assess their impact. 

\subsection{Baselines and competitors}~\label{sec:exp:baseline}
\noindent We integrate our proposed \sysname{} into the following two baseline frameworks: NeuS~\cite{wang2021neus} and Neuralangelo~\cite{li2023neuralangelo}.

\noindent \textbf{NeuS~\cite{wang2021neus}:}~NeuS proposes representing a surface as the zero-level set of a signed distance function (SDF) and introduces a new volume rendering method to train a neural SDF representation. By addressing inherent geometric errors, NeuS achieves accurate surface reconstruction, outperforming other methods for complex structures and self-occlusion.

\noindent \textbf{Neuralangelo~\cite{li2023neuralangelo}:}~Neuralangelo combines multi-resolution 3D hash grids and neural surface rendering to effectively recover detailed 3D surface structures from multi-view images, surpassing previous methods in fidelity and enabling large-scale scene reconstruction from RGB video captures. 

To validate the effectiveness of \sysname, we compare it to the following different competitors with the exact same input: pixelNeRF~\cite{yu2020pixelnerf}, MVSNeRF~\cite{sparesmvsnerf}, SparseNeuS~\cite{long2022sparseneus}, \textcolor{blue}{MonoSDF~\cite{yu2022monosdf}} , VolRecon~\cite{Ren_2023_CVPR}. 
pixelNeRF~\cite{yu2020pixelnerf} utilizes fully convolutional networks to effectively train on sparse view images and perform efficient view synthesis.
MVSNeRF~\cite{sparesmvsnerf} uses cost volumes from multi-view stereo to reconstruct neural radiance fields for physically based rendering.
SparseNeuS~\cite{long2022sparseneus} uses signed distance function and image features to enhance reconstruction quality through geometry reasoning and color blending.
\textcolor{blue}{MonoSDF~\cite{yu2022monosdf} integrates monocular depth and normal cues from pretrained predictors into neural implicit surface optimization to improve multi-view 3D reconstruction quality, especially in sparse view scenes.}
VolRecon~\cite{Ren_2023_CVPR} is the leading technique for detailed and noise-free scene reconstruction, utilizing SRDF and multi-view integration. {These methods are tailored for sparse reconstruction tasks, making a detailed comparative analysis among them both scientifically valid and justified}

Our network initialization scheme is similar to that of IDR~\cite{yariv2020multiview}. 
Our code is implemented using NeuS ~\cite{wang2021neus} and Neuralangelo ~\cite{li2023neuralangelo}, with the assumption that the region of interest is situated inside a unit sphere.

\subsection{Implementation Details}\label{sec:exp:implement}
\subsubsection{Training details}
We implement our \sysname{} on NeuS~\cite{wang2021neus} and Neuralangelo~\cite{li2023neuralangelo}.
For the NeuS+\sysname{} setting, we sample $512$ rays per batch and train our model for $300k$ iterations for all the experiments.
For the Neuralangelo+\sysname{} setting, we follow the original paper via setting the hash encoding resolution spans $2^5$ to $2^{11}$ with $16$ levels and train our \sysname{} with $500k$ iterations. 
All the experiments are running on a NVIDIA RTX 3090 GPU.
Training the NeuS+\sysname{} setting takes about 8 hours and Training the Neuralangelo+\sysname{} setting takes about 15 hours.
We train \sysname{} using the ADAM optimizer with a weight decay of $10^{-2}$ and a minium learning rate of $2.5\times10^{-5}$.
The $m$ and $n$ in \sysname{} is set to be $20$ and $768$ respectively.
For the semantic matching prior integration, we utilize a singler MLP mapping from the last dimension $768$ to $10$. 
We utilize the marching cubes algorithm to get a triangular meshes, with a resolution of $1024$ for all the tasks.

\subsubsection{Data preparation}
Similar to SparseNeuS~\cite{long2022sparseneus}, we employed background filtering techniques for all DTU, blendedmvs, and in-the-wild scenes.
In contrast to the intricate image processing techniques that SparseNeuS utilized, we adopted the Segment Anything Model~\cite{Kirillov_2023_ICCV} to effortlessly and efficiently acquire object masks for filtering during the training process.

\subsubsection{Network details}
Our semantic matching features are seamlessly integrated into the neural network architecture of the color network by employing a single-layer MLP with a size of $256\times7$. 
This integration efficiently reduces the dimensionality of high-dimensional semantic matching features, thus ensuring training stability without affecting the dimensions of other input.
The original encoder from Vision Transformer, with an output feature dimension of $768$, is utilized for extracting semantic patch features.
\color{blue}
\section{Experiment Results}\label{sec:results}

\begin{table*}[!ht]
\centering
\vspace{-0pt}
\setlength{\tabcolsep}{4.0pt}
\small
\scalebox{1.0}{
\begin{tabular}{l | ccccccccccccccc | c}
\toprule
Scan & 24 & 37 &  40 & 55 &63 &65& 69 &83 &97 &105& 106& 110& 114& 118& 122
  & Average \\
\midrule
COLMAP\cite{sfm}$^{\dag}$ &0.90 &2.89& 1.63& 1.08 &2.18 &1.94 &1.61 &\underline{1.30} &2.34& 1.28& \underline{1.10}& 1.42 &0.76& \underline{1.17} &\underline{1.14}&1.52 \\
\midrule
pixelNeRF\cite{yu2020pixelnerf} &5.13 &8.07 &5.85 &4.40 &7.11 &4.64 &5.68 &6.76 & 9.05  & 6.11 &3.95  &5.92  &6.26  &6.89  &6.93  & 6.18\\

MVSNeRF\cite{sparesmvsnerf} &1.96 &3.27  &2.54 &1.93  &2.57 &2.71 &1.82 &1.72 & 2.29  &1.75 & 1.72 &1.47  &1.29  &2.09  &2.26  &2.09 \\

SparseNeuS\cite{long2022sparseneus} &2.17 &3.29 &2.74 &1.67  &2.69  &2.42 & 1.58& 1.86& 1.94 &  1.35&1.50  &  1.45&  0.98& 1.86 & 1.87 &  1.96\\
\textcolor{blue}{MonoSDF\cite{yu2022monosdf}}   
& \textcolor{blue}{2.85} &  \textcolor{blue}{3.91} &\textcolor{blue}{2.26} & \textcolor{blue}{1.22} & \textcolor{blue}{3.37}& \textcolor{blue}{1.95} & \textcolor{blue}{1.95} & \textcolor{blue}{5.53}&  \textcolor{blue}{5.77} & \textcolor{blue}{1.10} & \textcolor{blue}{5.99} & \textcolor{blue}{2.28}& \textcolor{blue}{0.65} & \textcolor{blue}{2.65}& \textcolor{blue}{2.44}&  \textcolor{blue}{2.93}\\
VolRecon\cite{Ren_2023_CVPR}   &\underline{1.20}  &2.59 &1.56 &1.08 &1.43  & 1.92&1.11  & 1.48 & \underline{1.42} &1.05  & 1.19 &1.38  &  \underline{0.74}& 1.23 &1.27  & 1.38\\

\midrule
NeuS\cite{wang2021neus}  &4.57& 4.49 &3.97 &4.32 &4.63 &1.95& 4.68& 3.83& 4.15& 2.50& 1.52& 6.47& 1.26& 5.57& 6.11& 4.00
 \\

\sysname +NeuS &1.44 &\underline{2.14} &\underline{1.04} &\textbf{0.46} &\underline{1.22} &\underline{1.02} &\underline{1.01} & 1.64& 1.79 & \underline{0.94}& 1.37 & \underline{1.28} &  0.77& 1.19 &1.32 &  \underline{1.24}
 \\ 
 \midrule
Neuralangelo\cite{li2023neuralangelo} & 5.23& 4.37& 4.02 & 3.36& 3.27&2.53 &3.17 &3.02 &4.37  &2.43 & 1.49 &5.13  &1.37  & 4.20 & 4.79 & 3.52 \\

\sysname +Neuralangelo  & \textbf{1.07} & \textbf{2.03} &\textbf{0.97} & \underline{0.72}& \textbf{1.13}&\textbf{0.91} & \textbf{0.87}& \textbf{1.28}& \textbf{1.36} & \textbf{0.79} &\textbf{1.14}  &\textbf{1.23}  &\textbf{0.91}  &\textbf{1.11}  &\textbf{1.05}  &  \textbf{1.10}
 \\ 
\bottomrule

\end{tabular}}
\caption{%
Comparison with state-of-the-art alternatives with 3-views input on DTU benchmark. 
The bold denotes the best results, while the underlined indicates the second-best results. $^{\dag}$The results of COLMAP are based on standard library implementation.
}
\label{tab:DTUbenchmark}
\end{table*}

In this section, we conduct a series of experiments to comprehensively evaluate our proposed \sysname{}. We begin with quantitative comparisons against state-of-the-art methods on geometry reconstruction and novel view synthesis (\cref{sec:exp:quantative}). We then present extensive qualitative results on various datasets to demonstrate the visual superiority of our reconstructions (\cref{sec:exp:qualitative}). Finally, we perform detailed ablation studies and component analyses to validate the effectiveness of each part of our design (\cref{sec:exp:ablation}).
\color{black}

\subsection{Quantitative evaluation}~\label{sec:exp:quantative} 
Our quantitative results are presented in \cref{tab:DTUbenchmark}, employing the Chamfer distance metric, similar to the baseline~\cite{wang2021neus,li2023neuralangelo} and competitors.
This metric indicates the difference between the reconstruction geometry and the ground truth point clouds, with smaller values for better reconstruction.
This performance leap is further exemplified when integrating \sysname{} with baseline methods, yielding improvements of 2.76 with NeuS and 2.42 with Neuralangelo.
%
%
This suggests when the input viewpoints are severely limited (only $3$ viewpoints), the reconstruction accuracy of existing methods that are not specifically tailored for sparse reconstruction is significantly diminished.

Compared to traditional multi-view reconstruction methods like COLMAP~\cite{sfm}, our \sysname{} also perform much better by $0.42$.
The result of novel-view-synthesis is demonstrated in \cref{tab:novel view}.
\sysname{} boosts the baseline NeuS and Neuralangelo by $21.04\%$ and $8.72\%$ respectively.
This suggests that \sysname{} can also promote the performance for the novel view synthesis task.

\sysname{} showcases its superiority by surpassing methods that are solely designed for neural implicit reconstruction, sparse reconstruction, and traditional reconstruction approaches on sparse reconstruction tasks.
\sysname{} also boosts the baselines on novel view synthesis tasks consistently.
It highlights the effective performance of \sysname's components and their seamless integration as plugins.

\begin{table}[!ht]
\centering

\setlength{\tabcolsep}{6pt} 
\small
\scalebox{0.95}{  
\begin{tabular}{l|ccc}  
\toprule
Method & PSNR($\uparrow$) & LPIPS($\uparrow$) & SSIM($\downarrow$) \\
\midrule
Neus & 28.56 & 0.911 & 0.105 \\
Neus+\sysname & \underline{34.57} & \underline{0.986} & \underline{0.047} \\
Neuralangelo & 33.04 & 0.961 & 0.069 \\
Neuralangelo+\sysname & \textbf{35.92} & \textbf{0.989} & \textbf{0.041} \\
\bottomrule
\end{tabular}
}
\caption{
PSNR metric on novel view result.
\textbf{Bold} denotes best result and \underline{underline} denotes the second best result.
}
\label{tab:novel view}
\end{table}



\subsection{Qualitative results}~\label{sec:exp:qualitative}
We now present qualitative results to visually demonstrate the superiority of \sysname{}. We show comparisons on standard benchmarks, challenging real-world datasets, and for the novel view synthesis task. We have developed a project page to more effectively showcase our work:\href{https://seres0.github.io/}{https://seres0.github.io/}. We demonstrate various qualitative results from \cref{fig:DTUresult} to \cref{fig:wild}.

\subsubsection{DTU Dataset}
\begin{figure*}[!ht]
\centering
\includegraphics[width=1.00\linewidth]{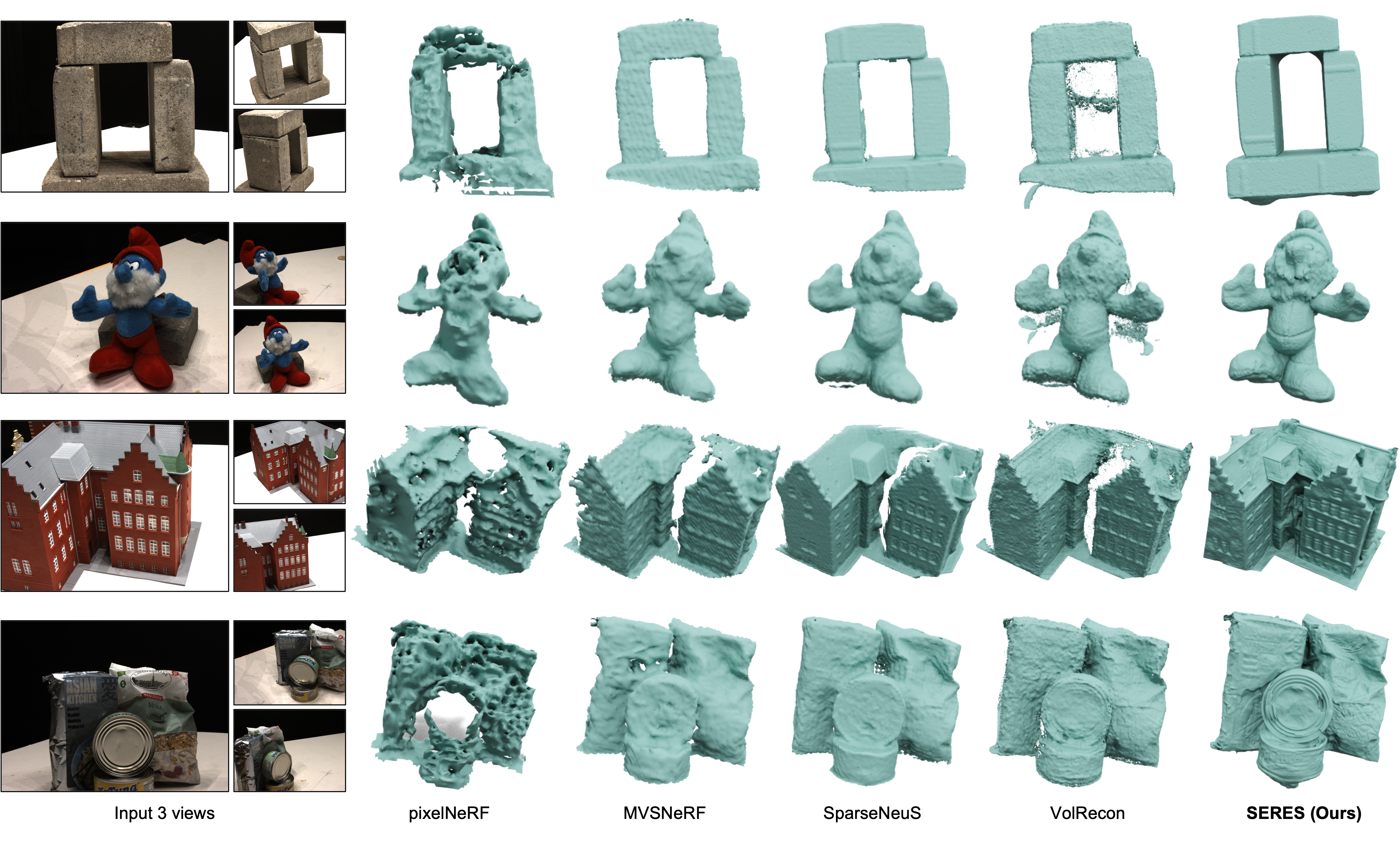}
\vspace{-20pt}
\caption{Results of reconstruction for models in the DTU~\cite{jensen2014large} benchmark from 3 sparse views (as shown in the first column).
The reconstruction results of prior methods are plagued by either geometric distortions (pixelNeRF~\cite{yu2020pixelnerf} and MVSNeRF~\cite{sparesmvsnerf}) or the loss of structural integrity (SparseNeuS~\cite{long2022sparseneus} and VolRecon~\cite{Ren_2023_CVPR}).
Our \sysname{} can reconstruct much more complete geometry with details. 
}\label{fig:DTUresult}

\end{figure*}

\begin{figure}[!ht]
\centering
\includegraphics[width=1.00\linewidth]{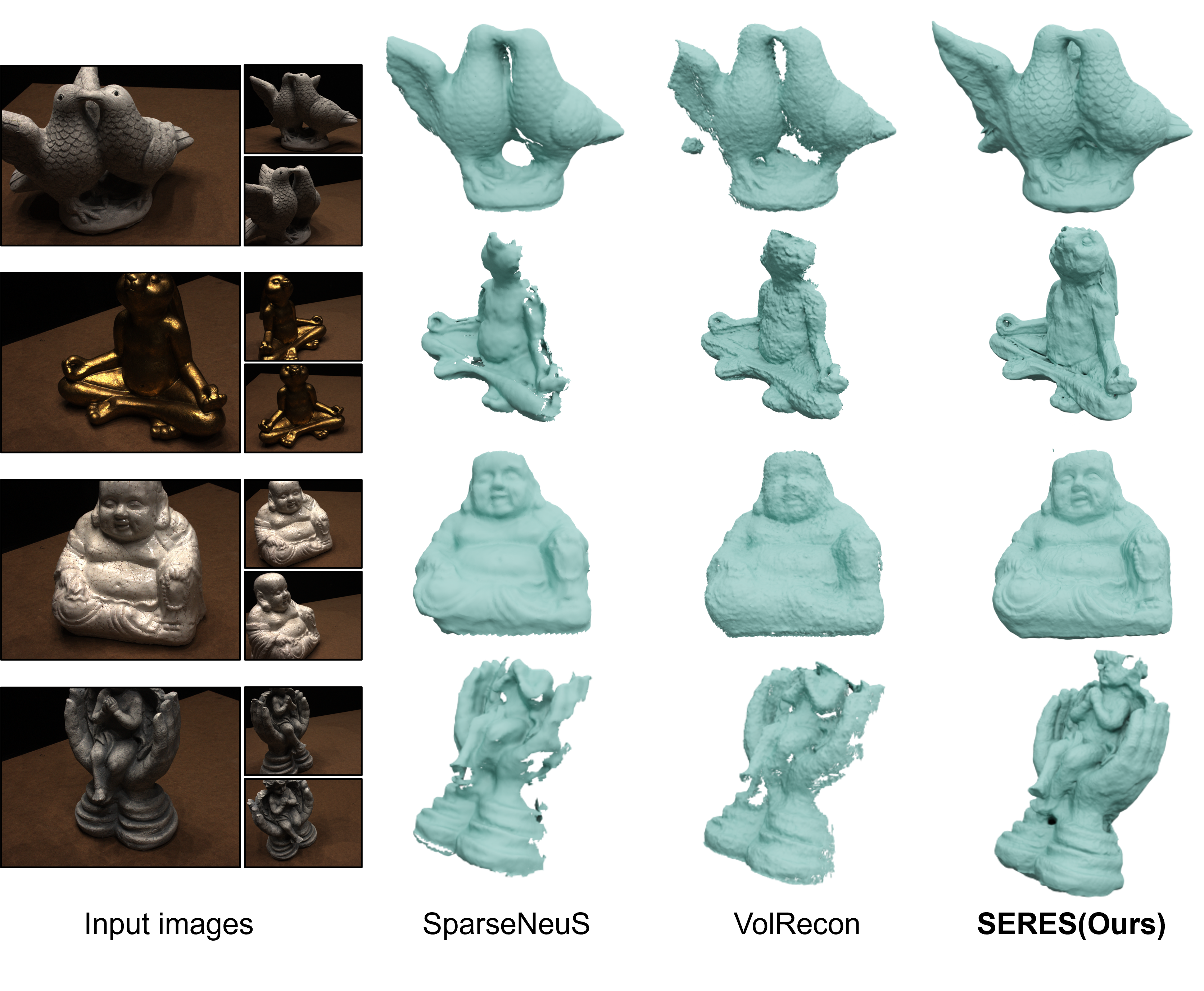}
\vspace{-18pt}
\caption{
%
Additional visual results of reconstruction for models in the DTU~\cite{jensen2014large} benchmark from 3 sparse views (as shown in the first column).
Our \sysname{} gives much more fine-grained and complete geometry structures when comparing to the SparseNeuS~\cite{long2022sparseneus} and VolRecon~\cite{Ren_2023_CVPR}.
}\label{fig:DTU_add}
\vspace{-5pt}
\end{figure}

~\cref{fig:DTUresult} gives the direct visualization of reconstructed geometry structures on DTU dataset.
The figure displays the $3$ input views, which are listed on the left side.
Previous methods fall short in reconstructing complete and fine-detailed geometry with only three input views.
The neural geometric implicit representations utilized by pixelNeRF~\cite{yu2020pixelnerf} and MVSNeRF~\cite{sparesmvsnerf} have not been effectively learned, resulting in overfitting that is particularly prominent in the input viewpoints. 
These input viewpoints are vulnerable to diverse geometric distortions.
While SparseNeuS~\cite{long2022sparseneus} and VolRecon~\cite{Ren_2023_CVPR} demonstrate a certain level of precision in structural learning, they display deficiencies in maintaining structural integrity.
Also, \cref{fig:DTU_add} illustrates additional results that the meshes extracted using VolRecon typically exhibit noisy surfaces. In contrast, our semantic-aware optimization approach yields results with greater detail and less noise. This improvement is largely due to VolRecon's reliance on volume representation, which lacks local surface constraints. When compared to SparseNeuS, our reconstructions are more complete and accurate. SparseNeuS tends to issue with radiance ambiguity, often resulting in incomplete and distorted geometries.

\subsubsection{BlendedMVS Dataset}
\begin{figure}[!ht]
\includegraphics[width=1.0\linewidth]{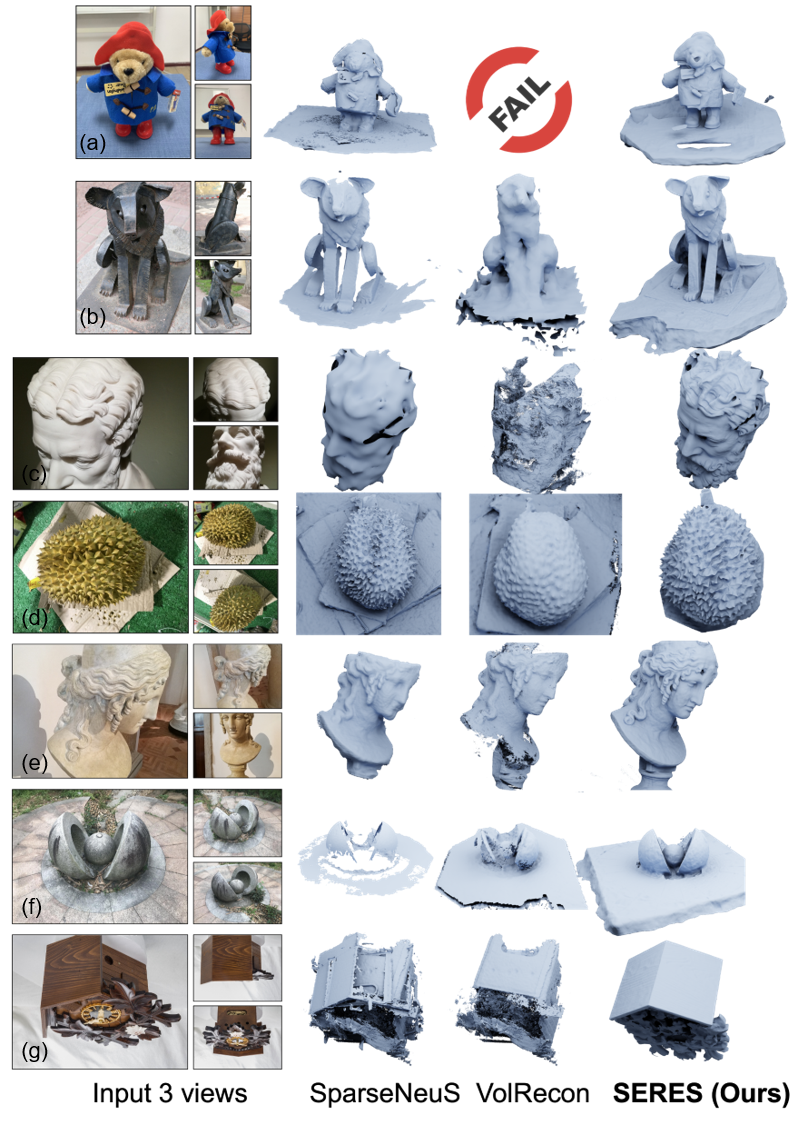}
\vspace{-18pt}
\captionsetup{labelfont={color=blue}, textfont={color=blue}}
\caption{\textcolor{blue}{
Our \sysname{} pipeline demonstrates superior performance in reconstructing models from the challenging BlendedMVS dataset\cite{yao2020blendedmvs} using only 3 sparse views as shown in the first column. Unlike prior methods that often result in incomplete structures with holes, our approach can reconstruct more complete geometry with details. 
}}\label{fig:bmvs}

\vspace{-5pt}
\end{figure}

\textcolor{blue}{We also evaluate our method on the challenging real-world dataset, BlendedMVS~\cite{yao2020blendedmvs}, with qualitative results presented in ~\cref{fig:bmvs}.  Since VolRecon~\cite{Ren_2023_CVPR} did not originally include results for BlendedMVS, we have reimplemented the model for this dataset. \sysname{} demonstrate strong generalizability, consistently producing cleaner and more complete results compared to both baselines~\cite{long2022sparseneus, Ren_2023_CVPR}. For instance, in the reconstruction of the dog (\cref{fig:bmvs}(b)), our model accurately captures its posture and geometric features. In contrast, VolRecon's global volume representation fails to render the complete object, while SparseNeuS produces a visibly broken geometry. Furthermore, our method excels in handling complex surface topologies, as highlighted by the intricate and successful reconstruction of the clock (\cref{fig:bmvs}(g)). These examples underscore the robust and superior reconstruction capabilities of our approach.}
Overall, SparseNeus fails to reconstruct accurate geometry due to large view difference inputs. VolRecon is not able to reconstruct the geometry faithfully. In contrast to this, our method can accurately and complementarily reconstruct complex geometry details owing to semantic-aware guidance. 

\subsubsection{Novel View Synthesis}

\begin{figure}[!ht]
\includegraphics[width=1.0\linewidth]{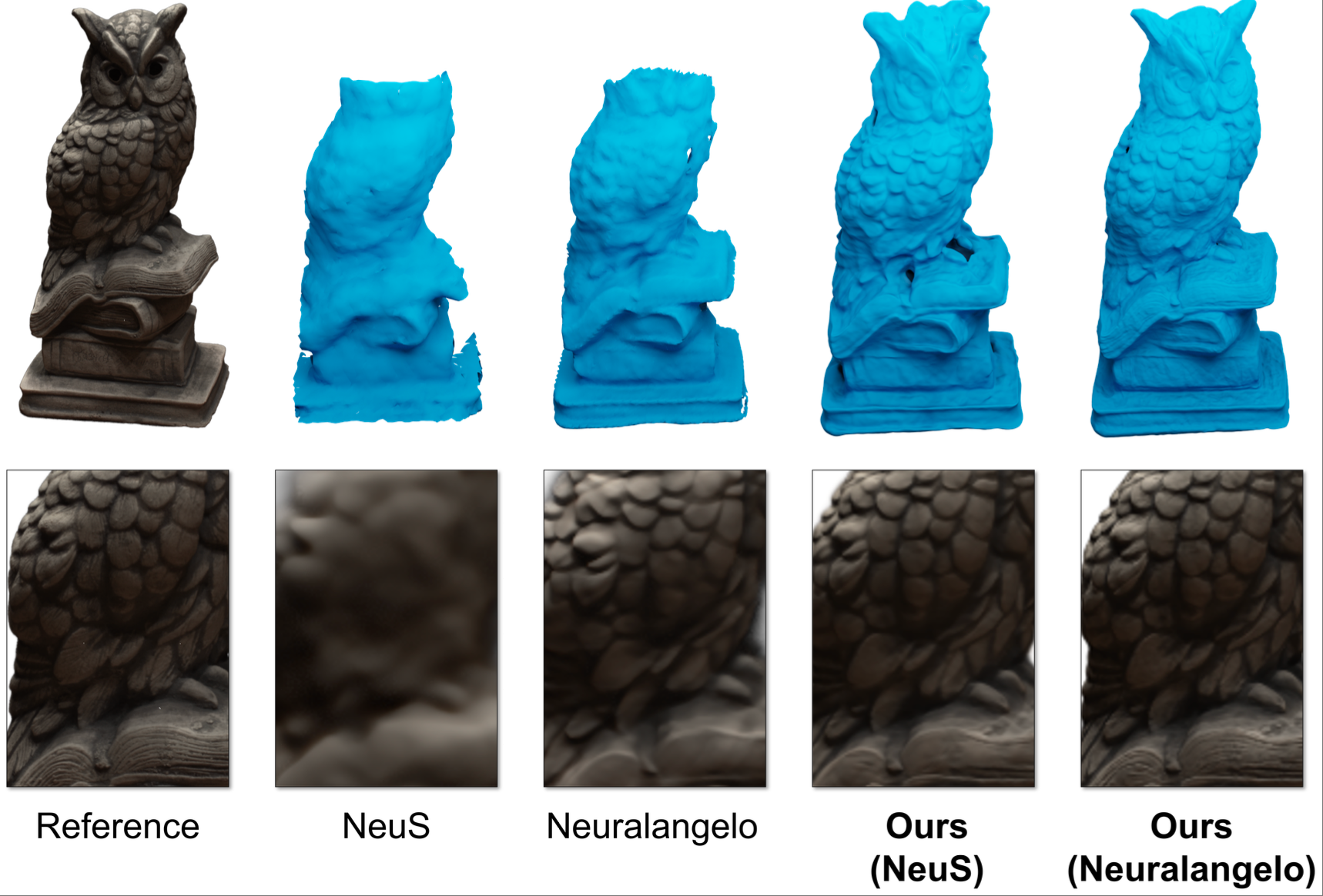}
\vspace{-18pt}
\caption{
Experimental tests of our \sysname{} as plugins to improve both the Neus and the Neuralangelo baselines, where the owl model is reconstructed from 3 views. It can be observed that quality of both the reconstructed geometry (top row) and the images obtained from novel view synthesis (bottom row) has been significantly improved to approach the ground truth (the first columen).
}\label{fig:owl}

\vspace{-5pt}
\end{figure}

\begin{figure*}[!ht]
\includegraphics[width=\linewidth]{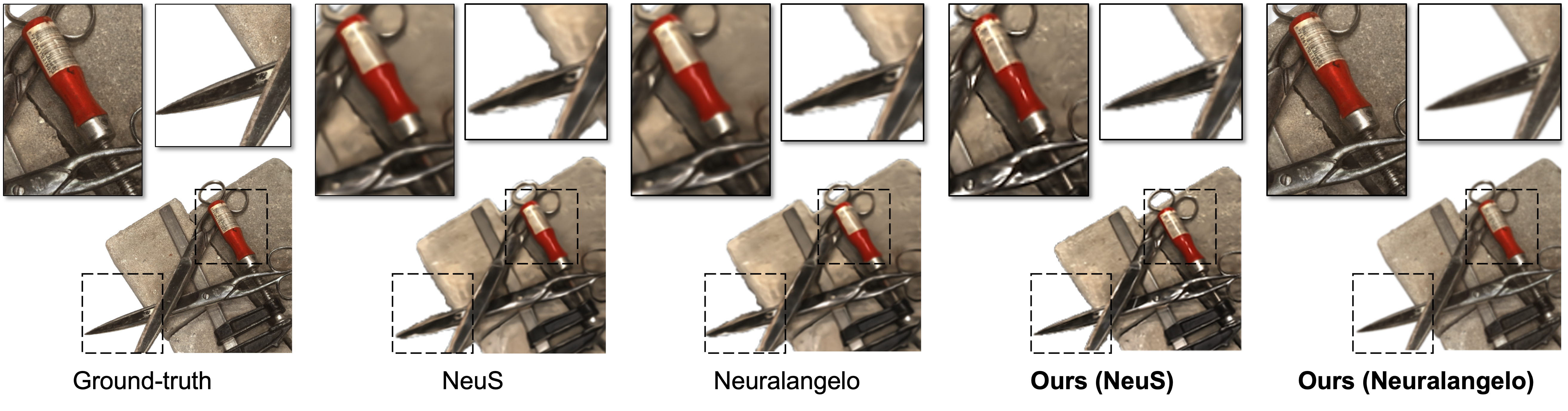}
\vspace{-20pt}
\caption{
Qualitative study of our method in the application of novel view synthesis by comparing to a few baselines and also the ground-truth -- it can be observed from the zoom-views that better fine-grained details can be reconstructed by our \sysname{} on both NeuS \cite{wang2021neus} and Neuralangelo \cite{li2023neuralangelo}. 
}\label{fig:nvs}

\end{figure*}

\cref{fig:owl} and \cref{fig:nvs} showcase the strength of our method, particularly in limited views. \sysname{} enhance novel view synthesis and neural implicit representations through our semantic-aware prior. When integrated as an add-on, our method significantly improves the performance of existing techniques like NeuS~\cite{wang2021neus} and Neuralangelo~\cite{li2023neuralangelo}. Our method markedly improves reconstruction by stitching up geometric information in areas that are challenging to match. In terms of novel view synthesis, it enhances clarity and detail, while effectively minimizing edge noise. 

\subsubsection{Wild Real-world Scenes}

\begin{figure*}[!ht]
\includegraphics[width=1.0\linewidth]{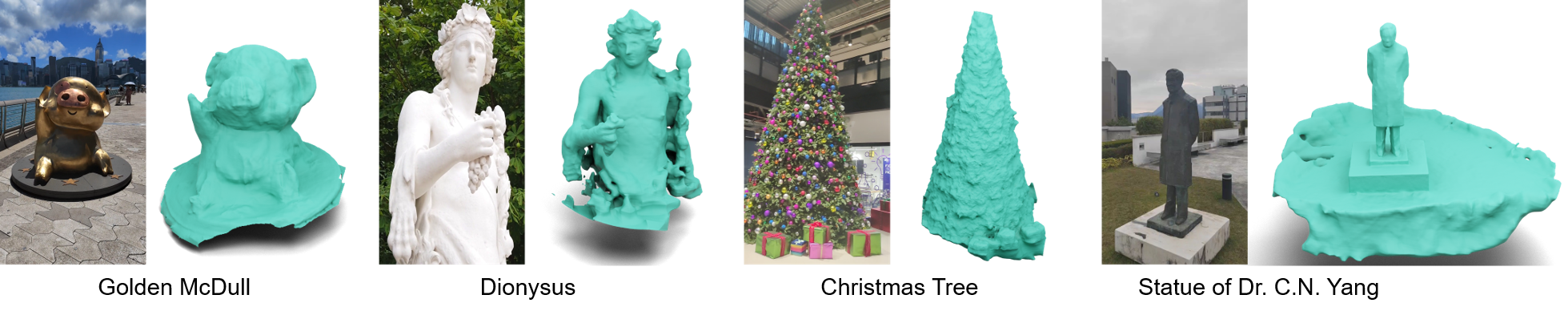}
\vspace{-18pt}
\captionsetup{labelfont={color=blue}, textfont={color=blue}}
\caption{
\textcolor{blue}{Our \sysname{} was deployed in four highly challenging in the wild scenes to investigate its performance.
These scenes exhibit the following characteristics: high reflection, complex structure, intricate topology, and lack of texture.
Even in these scenes with sparse input, \sysname{} is capable of reconstructing geometric structures that are approximately accurate.
}}\label{fig:wild}
\vspace{-5pt}
\end{figure*}

In \cref{fig:wild},~\sysname{} framework was tested in four highly challenging real-world scenes to evaluate its robustness and versatility. These scenes were characterized by high reflectivity, complex interactive structures, intricate topologies, and areas with minimal texture. Despite the sparse inputs (i.e, seven images for the Christmas tree, five each for the rest), \sysname{} successfully approximated the geometric structures in each scenario. The first scene featured a highly reflective Golden McDull, while the second scene showcased a statue with detailed, interwoven hand-held objects. The third scene involved a Christmas tree with complex topology and numerous thin structures. The final scene was a textureless, matte black statue. In all these diverse and challenging settings, our model consistently produced plausible and reasonable reconstructions.

\subsection{Ablation study and Analysis}\label{sec:exp:ablation}
\subsubsection{Ablation study}
\begin{figure*}[!ht]
\includegraphics[width=1.0\linewidth]{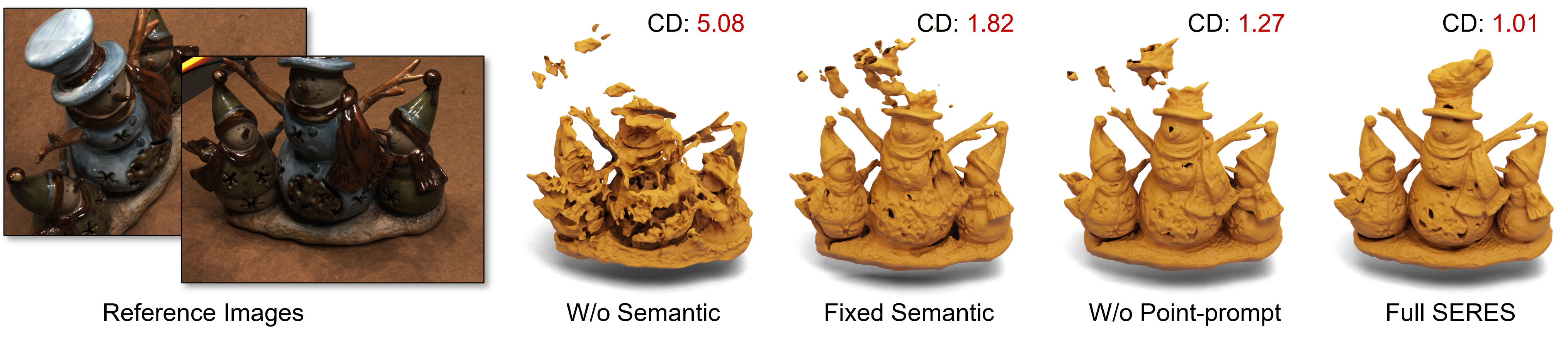}
    \vspace{-18pt}
    \caption{
    %
    %
Ablation study to explain the functionality of each module of our \sysname{} pipeline, where `W/o Semantic' and `Fixed Semantic' refer to the results without incorporating semantic priors vs giving a fixed semantic matching prior (i.e., without optimization). The results of `W/o Point-prompt' vs `Full' demonstrate that our point-prompt guided regularization effectively mitigates mesh noise and reduces shape ambiguity. The values of chamfer distance (CD) are also given for each results. 
    }\label{fig:ablation}
\end{figure*}

We systematically evaluate the impact of each component of our \sysname{} through an ablation study presented in \cref{fig:ablation}.
The comparison among ``W/o Semantic'', ``Fixed Semantic'', and ``Full'' suggests that integrating a correct semantic matching prior and then optimizing this prior helps the model to learn accurate geometry.
In contrast, when given an incorrect or fixed semantic matching prior, the model's performance will be significantly hindered.
Also, the reconstructed geometries of ``W/o Point-prompt'' and ``Full'' demonstrate that our point-prompt guided regularization effectively mitigates mesh noise and reduces shape ambiguity.
This demonstrates the effectiveness of all components in \sysname.

\color{blue}
\subsubsection{View-count analysis}
\begin{figure}[!ht]
\includegraphics[width=1.0\linewidth]{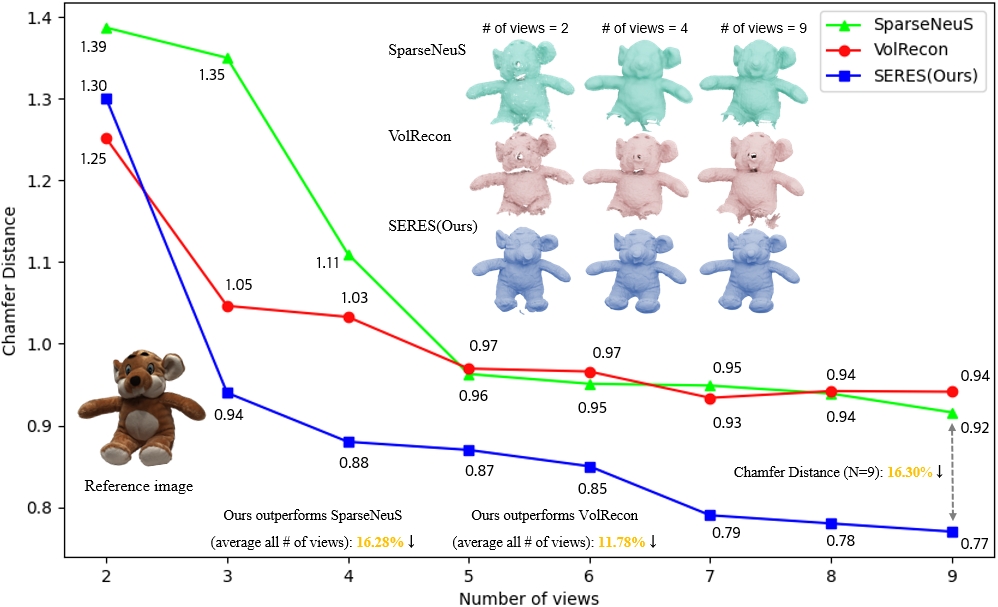}
\vspace{-15pt}
\caption{
This study demonstrates that using more number of views, ranging from $2$ to $9$, can reduce up to $40.8\%$ of the chamfer distances on the results of reconstruction. As can be found from the zoom-views, more input views are helpful to address radiance-ambiguity therefore contribute to the reconstruction of nuanced disparities. 
}\label{fig:number}
\end{figure}

\color{black}
We conducted an additional study examining the influence of the number of views on the reconstruction outcomes, as depicted in \cref{fig:number}.
As the number of inputs varies between $2$ and $9$, the chamfer distance exhibits a notable reduction of up to $40.8\%$. 
This reduction underscores the perceptible mitigation of shape ambiguity within the radiance field as the coverage of input views increases, which also demonstrates \sysname's strong performance.
Notably, our proposed \sysname{} demonstrates its capability to generate high-quality geometries even when presented with merely two views of an object, which demonstrates our superiority.

\color{blue}
\subsubsection{Point-prompt analysis}
\begin{figure}[!ht]
\includegraphics[width=1.0\linewidth]{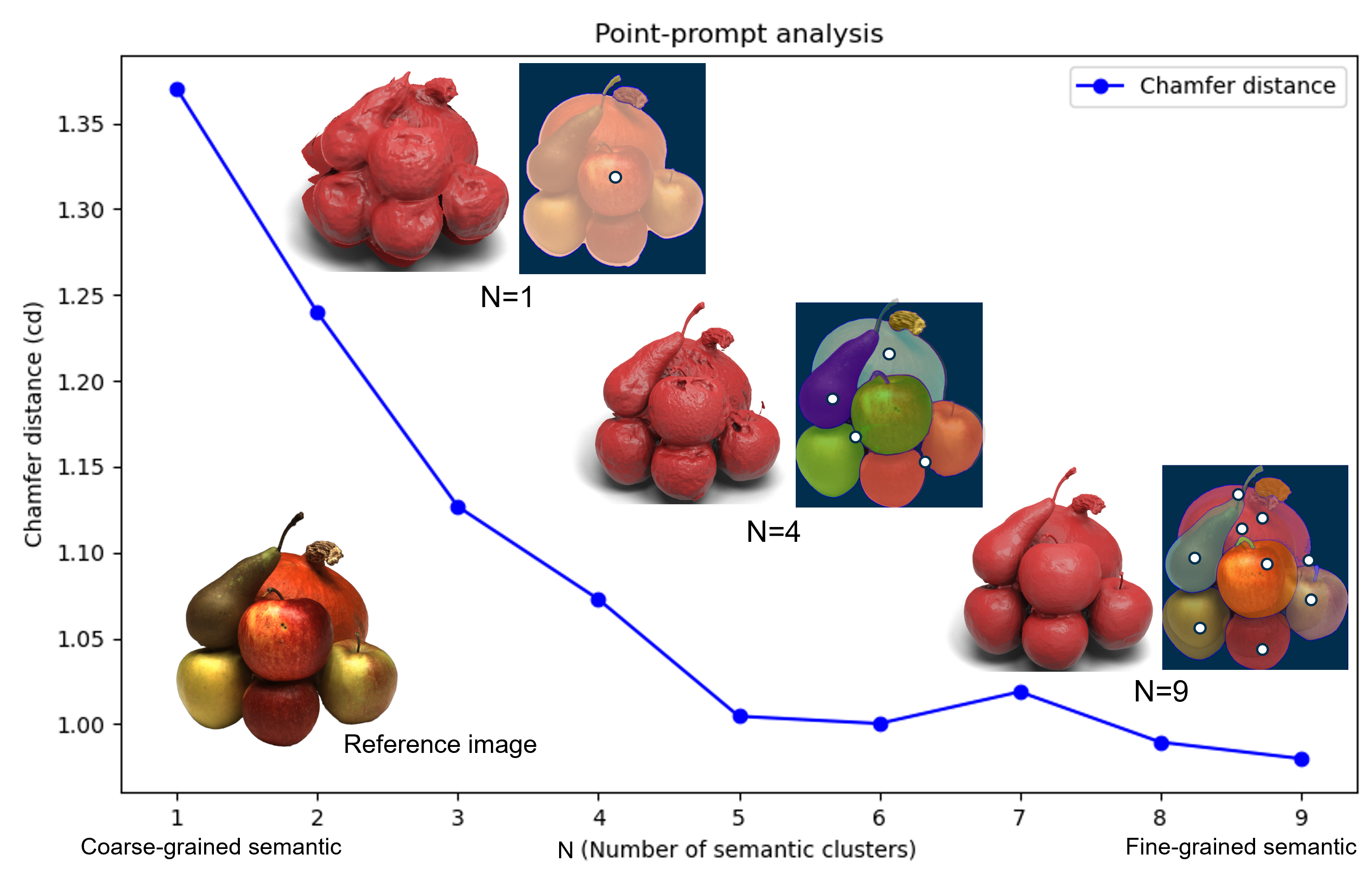}
\vspace{-18pt}
\caption{This figure illustrates the relationship between the chamfer distance and the number of semantic clusters.
The former metric is utilized for evaluating the quality of reconstruction, whereas the latter indicates the degree of semantic richness derived from point-prompts.
This figure simultaneously displays the reconstructed geometry and the masks of different categories obtained from the corresponding input point prompt coordinates ($N$=$1,4,9$).
}\label{fig:point_prompt}

\vspace{-12pt}
\end{figure}

\color{black}
We present a thorough analysis of the impact of the point-prompt guided regularization component in our \sysname{} on the reconstruction metric (Chamfer distance) in \cref{fig:point_prompt}.
The ~\cref{fig:point_prompt} suggests that as the value of $N$(number of clusters) increases from $1$ to $9$, the granularity of semantics becomes more refined (where distinct colors in the mask denote varying semantic categories). Simultaneously, there is a substantial enhancement in reconstruction quality.
The results demonstrate that the incorporation of masks with finer granularity derived from input point prompts significantly enhances the quality of reconstruction. 
Nevertheless, this trend gradually decelerates, indicating a diminishing incremental benefit from the addition of more point prompts.

\color{blue}
\subsubsection{Logits prediction}
\begin{figure}[!ht]
\includegraphics[width=1.0\linewidth]{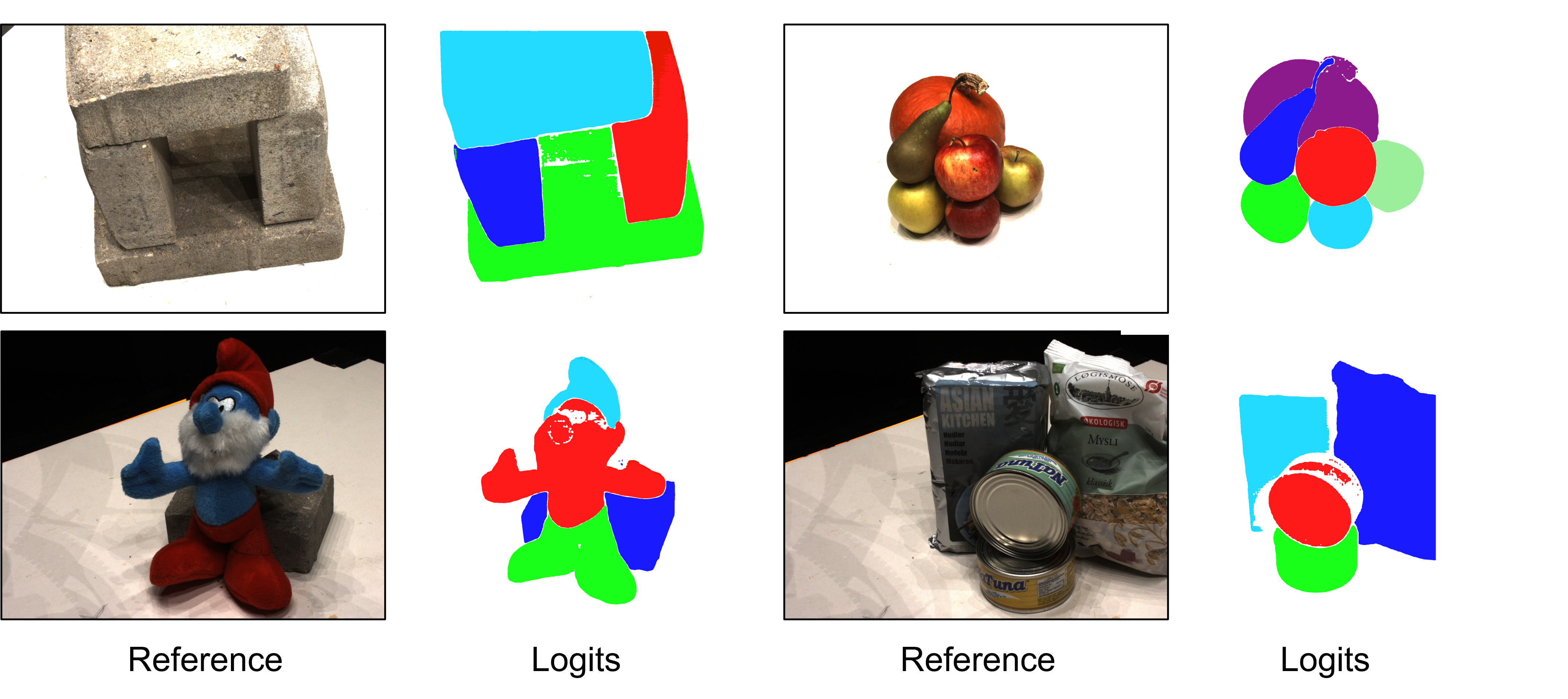}
\vspace{-18pt}
\caption{
This figure presents four sets of input view image reference and their corresponding semantic logits output for analysis.
Pixels of the same color indicate consistent categories defined by the highest logit value, whereas pixels of different colors signify varying categories associated with distinct maximum logit values.
}\label{fig:logit}

\vspace{-5pt}
\end{figure}

\color{black}
We demonstrate the visual result of logits prediction in \cref{fig:logit}.
The cases in \cref{fig:logit} are selected from the DTU dataset for demonstration.
The \cref{fig:logit} demonstrate that our semantic-aware neural field in \sysname{} successfully captures semantic categorical feature by employing point-prompt guided supervision and integrating semantic matching feature.
This confirms that our design is capable of learning semantic information, while also demonstrating that accurate semantic aids in geometric reconstruction.

\color{blue}
\subsubsection{Computational Cost}
To assess the practical viability of our method, we analyze its computational overhead.
As presented in \cref{tab:training_times}, incorporating \sysname{} as a plug-in module on top of the baseline results in only a marginal increase in training time, approximately 5\%. This indicates that our method can be integrated into existing pipelines without imposing a significant training burden.
Furthermore, we measure the average time required to extract prior information from a random set of input data, with details provided in \cref{tab:pre_time}. The entire process, including all constituent parts, completes in under 10 seconds.
These results collectively demonstrate that the performance gains afforded by \sysname{} are achieved with an acceptable and manageable computational cost, highlighting its efficiency and practicality.

\begin{table}[!ht]
\color{blue}
\centering
\setlength{\tabcolsep}{8pt}
\small
\begin{tabular}{l|c}
\toprule
\textbf{Baseline} & \textbf{Training Time (hours)} \\
\midrule
NeuS & 7.6 \\
SERES+NeuS & 8.0 \\
\midrule
Neuralangelo & 14.3 \\
SERES+Neuralangelo & 15.0 \\
\bottomrule
\end{tabular}
\captionsetup{labelfont={color=blue}, textfont={color=blue}}
\caption{Impact of SERES on training time. This table compares the training duration (in hours) on a single scene. Integrating SERES as a plug-in module results in only a marginal increase in training time, demonstrating its high computational efficiency. }
\label{tab:training_times}
\end{table}

\begin{table}[ht!]
\color{blue}
\centering
\setlength{\tabcolsep}{6pt} 
\small
\scalebox{0.95}{
\begin{tabular}{l|ccc}
\toprule
\textbf{Component} & SAM & ViT & Cross attention \\
\midrule
\textbf{Time(s)} & 8.750 & 0.280 & 0.004 \\
\bottomrule
\end{tabular}
}
\captionsetup{labelfont={color=blue}, textfont={color=blue}}
\caption{
Computational analysis of the semantic prior generation pipeline. This table details the average per-component inference time on the DTU dataset (for the default 3-input setting). The entire process is highly efficient, highlighting that the powerful semantic priors are obtained with only a minor computational cost at inference time.
}
\label{tab:pre_time}
\end{table}

\begin{figure}[!ht]
\includegraphics[width=1.0\linewidth]{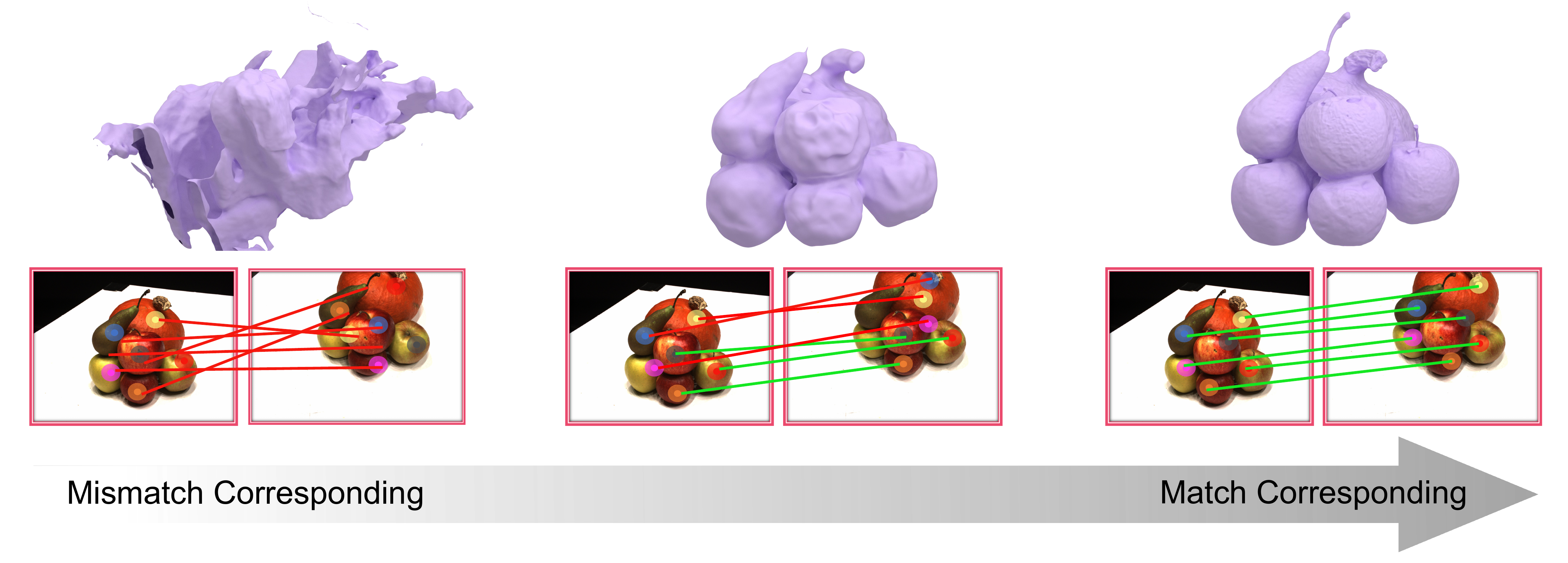}
\vspace{-18pt}
\captionsetup{labelfont={color=blue}, textfont={color=blue}}
\caption{\textcolor{blue}{
Visualization of trinocular view reconstruction results under three match accuracy levels (two input views are shown for clarity). From left to right: (a) completely random and out-of-order matches, (b) 50\% correct matches mixed with 50\% random/out-of-order matches, and (c) completely correct matches. The results highlight the critical role of match accuracy in determining reconstruction quality.
}}
\label{fig:corresponding}

\end{figure}

\subsubsection{Impact of Semantic Consistency}

To evaluate our model's robustness to semantic inconsistencies across views, we conducted a controlled experiment. We tested our pipeline with varying levels of semantic match accuracy between corresponding image patches, ranging from completely correct to entirely random. The specific accuracy levels and their visual results are detailed in \cref{fig:corresponding}.

The results reveal a clear correlation between match accuracy and reconstruction quality. With 100\% correct matches, our method produces high-quality geometry. Conversely, with completely random
and out-of-order matches, the reconstruction fails, confirming that meaningful semantic guidance is essential. Interestingly, with at least 50\% correct matches, the model produces a passable result; while noticeable degradation is present, the object's overall structure is still captured.

Crucially, our key finding is that the semantic matches generated by our default pipeline are themselves highly accurate, closely approaching the ideal 100\% correct match scenario. As a direct consequence, the resulting reconstruction quality is also excellent, far surpassing the degraded results of the 50\% accuracy case. It demonstrates that our approach effectively handles the minor inconsistencies typical of real-world data. This validates our method as both robust and practical, capable of achieving high-quality reconstruction without requiring perfectly curated semantic inputs.
\color{black}

\section {Limitations}
\color{blue}
While our method is robust to minor semantic inaccuracies, its performance degrades when the provided prior is fundamentally flawed. This limitation is demonstrated in \cref{fig:point_prompt}, where an inaccurate user prompt leads to a poor reconstruction.

Our method's primary limitation arises when reconstructing scenes with highly complex surface topologies. This challenge is evident in our results for the Christmas tree in \cref{fig:wild}. In this case, our method produces a plausible overall shape but fails to capture fine-grained details like individual branches.

This issue stems from two fundamental sources. The first is the well-known spectral bias of MLPs, which inherently struggle to model high-frequency geometric details. The second is the 2D nature of our semantic prior. The segmentation model (SAM) tends to group an entire complex object, such as the tree, into a single monolithic semantic region. This coarse prior lacks the detailed topological information required for precise reconstruction. As a result, it over-regularizes the geometry towards an overly smooth surface, reinforcing the MLP's low-frequency bias.


Our approach also inherits challenges common to all multi-view reconstruction methods. For instance, it struggles with highly reflective surfaces. This problem leads to visual artifacts and geometric distortions, as seen in the reconstructions of the high reflective Golden McDull (\cref{fig:wild}) and the metal rabbit (\cref{fig:DTU_add}). Such surfaces violate the multi-view photometric consistency assumption. Our semantic boundaries can mitigate this ill-posed problem but cannot fully resolve it.

Similarly, our method has difficulty faithfully reconstructing large textureless areas. The face of the statue of Dr. C.N. Yang (\cref{fig:wild}) is a clear example of this limitation. The issue is caused by severe photometric ambiguity. A lack of distinct visual features leaves the geometry under-constrained, even within the region defined by our semantic mask. These challenges highlight clear directions for future improvements.


\color{black}
\section{Conclusion}\label{sec:conclusion}
We presented \sysname{} in this paper as a novel semantic-aware framework to reconstruct 3D models from sparse views. By incorporating semantic-aware encoding and leveraging the segment anything model, our framework can effectively tackle the challenges of reconstructing 3D geometric details from a very few number of input views. The newly proposed point-prompt regularization enhances the neural implicit models, ensuring semantic region constraints and improving the representations of intricate object boundaries.
%
%
Extensive experiments on diverse datasets demonstrate the superiority of our \sysname{} over a variety of baseline methods, verifying its effectiveness in sparse view reconstruction.
%

%
%

\section*{Acknowledgments}
This work was supported by the National Natural Science Foundation of China under Grants 62422311 and 62176152. This work is partially supported by the Centre for Perceptual and Interactive Intelligence (CPII) Ltd., a CUHK-led under the InnoHK scheme of Innovation and Technology Commission.

\FloatBarrier


%





\ifCLASSOPTIONcaptionsoff
  \newpage
\fi

\bibliographystyle{IEEEtran}
\normalem
\bibliography{reference}

\end{document}